%% file: main.tex
\documentclass[runningheads]{llncs}

\usepackage[mobile]{eccv}

\usepackage{eccvabbrv}

\input{arx_preamble}

\usepackage[accsupp]{axessibility}  

\usepackage[breaklinks,colorlinks,citecolor=eccvblue]{hyperref}

\usepackage[capitalize]{cleveref}
\crefname{section}{Sec.}{Secs.}
\Crefname{section}{Section}{Sections}
\Crefname{table}{Table}{Tables}
\crefname{table}{Tab.}{Tabs.}

\begin{document}

\title{Freeview Sketching: View-Aware Fine-Grained Sketch-Based Image Retrieval}

\titlerunning{Freeview Sketching: View-Aware Fine-Grained Sketch-Based Image Retrieval}

\author{
\MYhref[eccvblue]{https://aneeshan95.github.io/}{Aneeshan Sain} \hspace{.1cm}
\MYhref[eccvblue]{http://www.pinakinathc.me/}{Pinaki Nath Chowdhury} \hspace{.1cm} 
\MYhref[eccvblue]{https://subhadeepkoley.github.io/}{Subhadeep Koley} \\
\MYhref[eccvblue]{https://ayankumarbhunia.github.io/}{Ayan Kumar Bhunia} \hspace{.1cm} 
\MYhref[eccvblue]{https://personalpages.surrey.ac.uk/y.song/}{Yi-Zhe Song}\vspace{-2mm}
}

\authorrunning{A.Sain et al.}
\institute{SketchX, CVSSP, University of Surrey, United Kingdom.
\\{\tt\small \{a.sain, p.chowdhury, s.koley, a.bhunia, y.song\}@surrey.ac.uk}
}

\maketitle

\begin{abstract}
\vspace*{-7mm}
In this paper, we delve into the intricate dynamics of Fine-Grained Sketch-Based Image Retrieval (FG-SBIR) by addressing a critical yet overlooked aspect -- the choice of viewpoint during sketch creation. Unlike photo systems that seamlessly handle diverse views through extensive datasets, sketch systems, with limited data collected from fixed perspectives, face challenges. Our pilot study, employing a pre-trained FG-SBIR model, highlights the system's struggle when query-sketches differ in viewpoint from target instances. Interestingly, a questionnaire however shows users desire autonomy, with a significant percentage favouring view-specific retrieval. To reconcile this, we advocate for a view-aware system, seamlessly accommodating both view-agnostic and view-specific tasks. Overcoming dataset limitations, our first contribution leverages multi-view 2D projections of 3D objects, instilling cross-modal view awareness. The second contribution introduces a customisable cross-modal feature through disentanglement, allowing effortless mode switching. Extensive experiments on standard datasets validate the effectiveness of our method.
\vspace*{-2mm}
\vspace*{-2mm}
\end{abstract}


\section{Introduction}
\vspace*{-2mm}

Sketch, a versatile medium, stands out as an exceptional input query modality, especially in the face of image retrieval. As a complement to text, it offers a distinct level of fine-grained expressiveness, making it a superior input modality~\cite{bhunia2021more} for fine-grained image retrieval. The past decade has witnessed extensive research in Fine-Grained Sketch-Based Image Retrieval (FG-SBIR) \cite{song2017deep,yu2016sketch,sain2023exploiting,bhunia2022adaptive}, delving into the unique characteristics of sketch data, such as abstraction \cite{umar2019goal}, style \cite{sain2020cross}, data scarcity \cite{bhunia2021more}, and drawing order \cite{bhunia2020onthefly}. However, in this paper, we take a departure from the intricacies of sketch-specific traits and shift our focus to a fundamental aspect of the human experience -- their interaction with the system. Specifically, we address a noteworthy challenge neglected thus far in the literature: ``Which view should I sketch?" -- a question that we hear a lot!

\begin{SCfigure}[50][h]
\includegraphics[width=0.62\linewidth]{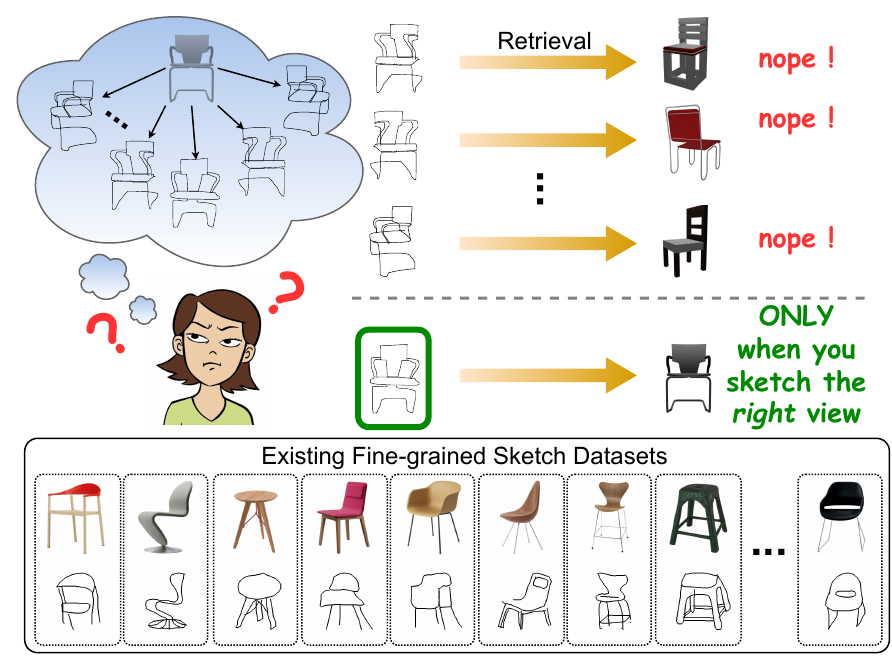}
\caption{While searching a specific photo, users are often confused as to `Which view should I sketch?' Adding to their worries, models trained on existing FG-SBIR datasets that have sketch-photo pairs matched against a \textit{fixed view}, fail to fetch a photo even when it's present in the gallery, unless its \textit{view matches} that of the query-sketch \textit{ exactly}. We aim to alleviate this issue and incorporate \textit{view-awareness} in the FG-SBIR paradigm.}
\label{fig:opener1}
\end{SCfigure}

The ``view'' predicament is inherently intuitive -- just as individuals carefully choose optimal camera angles when capturing photos, they also deliberate on the best view to portray an object before sketching \cite{qi2021FGSBSR}. In contemporary photo-based systems, this view problem is typically addressed in a data-driven manner, leveraging extensive image datasets to ensure comprehensive coverage of various object views, essentially making them view-invariant \cite{monnier2022share}. However, this seamless solution does not extend to sketches. The constrained nature of available sketch data, often collected from fixed viewpoints, introduces a significant bias toward these limited perspectives. Consequently, if your input sketch is executed from an unintended view, the system will respond less forgivingly (\cref{fig:opener1}). 

\vspace{-7mm}
\sidecaptionvpos{figure}{c}
\vspace{-8mm}

This assertion is validated through our pilot study. We employed a FG-SBIR model pre-trained on existing fine-grained sketch datasets, featuring single-view matched sketch-photo pairs. The model was then evaluated on a carefully curated gallery set where query-sketches deliberately did not align with the view of their corresponding target instances. As anticipated, the baseline FG-SBIR model faced substantial challenges, with a majority of retrieval attempts failing to identify the correct target instance (drops top-1 \cite{yu2016sketch} accuracy by $27.15$\%). Upon scrutinising the results, a clear pattern emerged -- incorrect photos retrieved at \textit{rank-1} were not only structurally similar but, more crucially, shared the same view as that of the query-sketch, echoing with earlier findings in the field \cite{sain2023clip,pang2020solving}. 

A user experience questionnaire however revealed somewhat contrasting conclusions. While the baseline FG-SBIR \cite{yu2016sketch} model exhibited a bias towards shape-matching \cite{bhunia2022adaptive}, users expressed a desire for autonomy within the system. A notable percentage (64.56\%) of participants, particularly those adept at sketching, indicated a preference for view-specificity as a feature when using sketches for retrieval. Essentially, they articulated a desire for the system to precisely retrieve what they had sketched, aligning with their specific viewpoint preferences.

\begin{wrapfigure}[13]{l}{0.6\linewidth}
    \centering
    \vspace*{-8mm}
    \includegraphics[width=\linewidth]{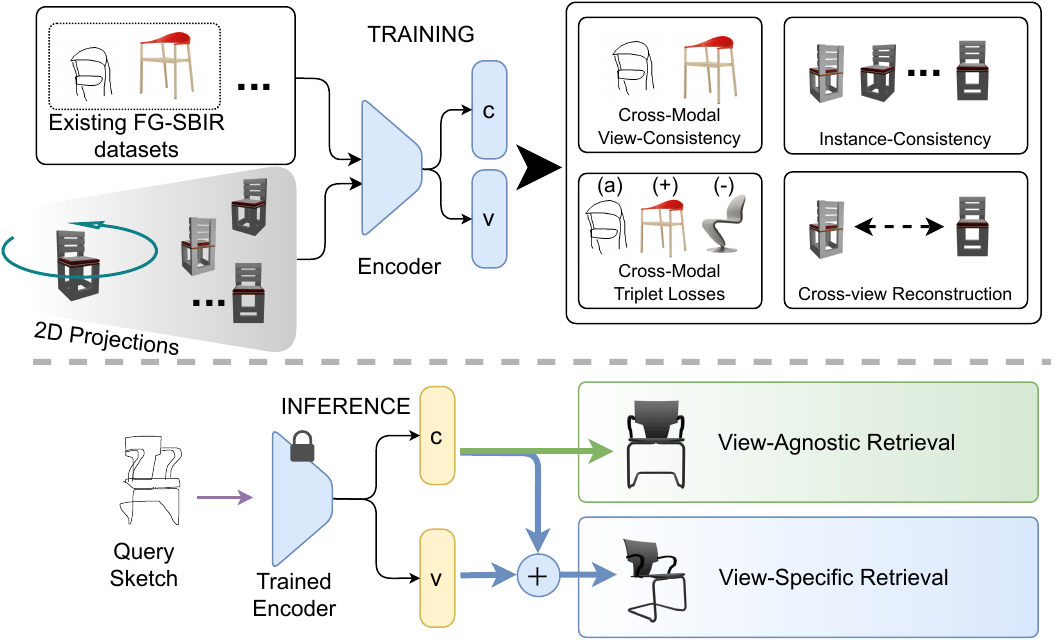}
    \vspace*{-7mm}
    \caption{Our framework. We aim to handle both view-\textit{agnostic} and \textit{specific} retrieval using one model.}
    \vspace*{-2mm}
    \label{fig:opener2}
\end{wrapfigure}

Our approach to addressing the ``view'' problem therefore does not lean towards creating a system that completely ignores views (view-agnostic) or one that is strictly sensitive to view changes (view-specific). Instead, we advocate for a view-aware system. In other words, we aim for a system that can seamlessly adapt to both scenarios simultaneously. We envision a system capable of handling both view-agnostic and view-specific tasks (see \cref{fig:opener2}) without requiring any redesign or additional training -- simply flipping a switch should suffice! 

Establishing view-awareness poses a substantial challenge, primarily due to the limitations inherent in existing fine-grained datasets: \textit{(i)} absence of view-specific annotations or information in sketch-photo pairs crucial for developing view-awareness, and \textit{(ii)} For every sketch-photo pair, sketch is created \cite{qi2021FGSBSR} from a fixed single point of view, reflecting structural matching against a single photo captured from that specific perspective. As our first contribution, we aim to alleviate this by leveraging sketch-independent multi-view 2D rendered projections of 3D objects \cite{shapenet}, to gain view-aware knowledge and associate it with the cross-modal sketch-photo discriminative knowledge learnt using standard FG-SBIR datasets \cite{yu2016sketch,sangkloy2016sketchy}, thus distilling cross-modal view awareness into the pipeline. 

Our second contribution revolves around designing a cross-modal feature that is customisable for both view-agnostic and view-specific retrieval simultaneously. To achieve this, we adopt a feature disentanglement framework commonly found in the literature \cite{wang2018orthogonal, sain2021stylemeup,yang2019disentangling}. In this framework, we disentangle sketch features into two distinct parts: content and view. The content part encodes the semantics present in the sketch, while the view part encapsulates view-specific features. At inference time, the key decision lies in selecting which part(s) of the feature to utilise -- choosing only the content results in view-agnostic retrieval, whereas combining content and view retains view specificity. Essentially, it is a flip of a switch! While this sounds straightforward, implementing it is non-trivial. To facilitate this disentanglement, we introduce two specific designs, particularly crucial in the absence of ample sketch view data. 

Concretely speaking, first, we enforce instance-consistency across multi-view
2D rendered projections of a 3D model (\cref{fig:opener2}) by constraining their disentangled \textit{content} parts to sit together, and also introduce a cross-view reconstruction objective, which merges \textit{content} of one projection with \textit{view} of another to reconstruct the latter. Second, we impose a view-consistency objective between a \textit{matching} sketch-photo pair, by constraining their \textit{view} parts to be closer in the embedding space, thus ensuring the associativity of paired sketch-photo views.

In summary, \textit{(i)} we propose a view-aware system designed to address the often-overlooked challenge of choosing the appropriate view for sketching, accommodating both view-agnostic and view-specific retrieval seamlessly. \textit{(ii)} we introduce the use of multi-view 2D rendered projections of 3D objects to overcome the limitations of existing datasets, promoting cross-modal view awareness in the FG-SBIR pipeline. \textit{(iii)} we present a customisable cross-modal feature through a disentanglement framework, allowing users to effortlessly switch between view-agnostic and view-specific retrieval modes.

\vspace*{-5mm}
\section{Related Works}
\vspace*{-3mm}

\noindent \textbf{Fine-Grained SBIR:} 
Although Sketch-based Image Retrieval (SBIR) began as a category-level task \cite{sangkloy2016sketchy, collomosse2017sketching,sain2022sketch3t}, research quickly shifted to fine-grained SBIR where, the aim is to retrieve the only matching photo from a gallery of same-category photos, with respect to a query-sketch. Starting from deformable part models~\cite{li2014fine}, numerous deep approaches have emerged \cite{yu2016sketch,song2017deep,pang2019generalising,bhunia2021more,sain2020cross}, centering around a triplet-loss based deep Siamese networks~\cite{yu2016sketch} that learn a joint sketch-photo manifold. Encouraged by new datasets with fine-grained sketch-photo association \cite{yu2016sketch,sangkloy2016sketchy, chowdhury2022fs} FG-SBIR improved further with hybrid cross-domain image generation \cite{pang2017cross}, textual tags \cite{song2017fine}, local feature alignment strategy \cite{xu2021dla}, attention mechanism involving higher-order~\cite{song2017deep} or auxiliary losses \cite{lin2019tc},  and mixed-modal jigsaw solving based pre-training strategy \cite{pang2020solving}, to name a few. Apart from addressing specific sketch-traits, like hierarchy of details \cite{sain2020cross}, style-diversity \cite{sain2021stylemeup}, or redundancy of strokes \cite{bhunia2022sketching}, works explored various application scenarios like cross-category generalisation \cite{pang2019generalising, bhunia2022adaptive}, overcoming data-scarcity \cite{sain2023exploiting, bhunia2021more}, early-retrieval \cite{bhunia2020onthefly},  and recently zero-shot cross-category FG-SBIR \cite{sain2023clip}. Contemporary research has extended FG-SBIR to scene-level retrieval, modelling cross-modal region associativity \cite{chowdhury2022partially} and enhanced further using text as an optional query \cite{chowdhury2023scenetrilogy}. 
These works however, have largely ignored the question of `view-awareness' in context of FG-SBIR. In this work, we thus aim to incorporate `view-awareness' in FG-SBIR as two branches namely view-\textit{agnostic} FG-SBIR and view-\textit{specific} FG-SBIR.

\noindent \textbf{Learning 3D knowledge from 2D images}
The concept of view-awareness arises from the underlying motivation of representing an object holistically from all view-perspectives. In this regard numerous works have attempted at understanding the task of 3D shape retrieval \cite{klokov2017escape,he2018retrieval,wang2017cnn}. 3D shape retrieval methods can be broadly divided as 3D Model-based methods that directly learn shape features from 3D data formats like polygon meshes \cite{boscaini2016learning,xie2016deepshape,zhu2016learning}, voxels \cite{li2016fpnn,qi2016volumetric,sedaghat2016orientation}, and  point-clouds \cite{wu20153d,qi2017pointnet}; and view-based methods \cite{su2015recognition}.
While earlier works applied view based similarity against pre-processed 3D shape descriptors \cite{funkhouser2003search}, to retrieve 3D models using a 2D image, others encouraged lesser views by clustering \cite{cyr20013d}.
Recent improvements include real-time 3D shape search engines based on the 2D projections of 3D shapes \cite{bai2016gift} or using max-pooling to aggregate features of different views from a shared CNN like MVCNN \cite{su2015recognition}.
%
Limited by the availability of multiple views at large-scale, single-view 3D shape learning had gained traction. While SSMP \cite{ye2021shelf} uses adversarial regularisation towards shape learning it often falls unstable for complex structures. Others include semantic regularisation for implicit shape learning \cite{huang2022planes}, or a 3-step learning paradigm \cite{alwala2022pre} for scalable shape learning including synthetic data pre-training. However most of such methods are focused on images alone unlike our cross-modal retrieval setup. Moreover, curating multi-view photos of an object is relatively easier than our case of collecting multi-view sketches -- almost all FG-SBIR datasets \cite{yu2016sketch,sangkloy2016sketchy} contain only one photo matched against a sketch, from a fixed view. Furthermore sketch being quite sparse and lacking visual cues, aligning it with a 3D model is in itself quite challenging. In light of such limitations, we advocate for a simpler strategy without training on 3D descriptors, and relying only on 2D views of unpaired photos to instill view-awareness in FG-SBIR.

\noindent \textbf{Sketch-Based 3D Shape Retrieval:} Being similar to FG-SBIR in using sketch as a fine-grained query, we explore a few relevant works on this well explored sketch-related 3D application, where the aim is to retrieve 3D shapes given a sketch query. While prior studies primarily focused on category-based retrieval, aiming to retrieve shapes of the same category as that of the query sketch \cite{yoon2017retrieval, saavedra2012retrieval, li2013retrieval, li2017retrieval},
recent deep learning methods attended to mapping sketch and shape-features to a common shared embedding space~\cite{wang2015retrieval, xu2016retrieval, dai2017retrieval, xie2017retrieval, he2018retrieval, su2015recognition}. A major concern similar to ours in this regard however, is the issue of view-variance \ie, multiple sketches can be drawn from various views of one 3D shape. To alleviate the same, earlier works encoded rendered 2D projections \cite{xu2022retrieval} via CNNs \cite{su2015recognition}, as Wasserstein barycenters \cite{xie2017retrieval}, or used triplet-center loss \cite{he2018retrieval}. 
Others include learning intra and cross-domain similarities from just two views \cite{wang2015retrieval}, or computing global 3D-shape descriptors \cite{daromLDSHIFT, dai2017retrieval}. 
Shifting to fine-grained paradigm further intensifies challenges owing to a sketch's unconstrained free-hand deformations \cite{sain2021stylemeup} and lack of large-scale datasets \cite{sherc22}. While \cite{qi2021FGSBSR} approaches by creating a dataset of 4,680 sketch-3D pairs with multiple 2D shape projections for improved retrieval, \cite{chowdhury2023democratising} learns the correspondence between a set of 3D points and their 2D projections. In such settings however learning/optimising over 3D shape data is mostly pivotal to the training procedure. Our motivation however is to keep the training paradigm restricted in 2D for simplicity, -- use only the 2D projections of an object while accessing only one single-view sketch per instance as available in existing FG-SBIR datasets. 

\vspace*{-4.5mm}
\section{Problem and Analysis}

\vspace*{-2mm}
\subsection{Background on FG-SBIR} \label{sec:fgsbir_back}
\vspace*{-1mm}

Given a query-sketch ($s$), FG-SBIR \cite{yu2016sketch} refers to the task of retrieving a particular matching instance from a gallery of photos of the same category, where the underlying convention is to associate \textbf{one 2D photo per sketch}. This convention is followed in standard FG-SBIR datasets like QMUL-ChairV2 \cite{yu2016sketch}, QMUL-ShoeV2 \cite{yu2016sketch} and Sketchy \cite{sangkloy2016sketchy} that comprises $k$ instance-level sketch/photo pairs as $\{s_i, p_i\}_{i=1}^{k}$ (per category for Sketchy \cite{sangkloy2016sketchy}), where the sketches are drawn from a fixed view corresponding to the object-photo. 
A baseline FG-SBIR framework \cite{yu2016sketch} usually aims to learn an embedding function 
$\mathcal{E}_\theta: \mathcal{I} \rightarrow \mathbb{R}^d$
that maps a rasteried sketch/photo, $\mathcal{I} \in \mathbb{R}^{H\times W\times 3}$ to a d-dimensional feature $f_\mathcal{I} \in \mathbb{R}^d$.
$\mathcal{E}_\theta(\cdot)$ is usually a CNN \cite{yu2016sketch, song2017deep} or Transformer \cite{sain2023exploiting} based encoder, that is shared between sketch and photo branches, and trained over a triplet-loss based objective \cite{yu2016sketch} ($\mathcal{L}_\text{Tri}$) where the distance $\delta(a,b)$ = $||a-b||_2$ between features of query-sketch ($f_s$) and its matching photo ($f_p$) is reduced, while increasing it from a random photo-feature $(f_n)$ in the joint sketch-photo embedding space, as:
\vspace*{-2mm}
\begin{equation}
\label{equ:triplet_basic}
    \mathcal{L}_{\text{Tri}} = \max \{0, \mu + \delta(f_s, f_p) - \delta(f_s, f_n) \} \;,
    \vspace*{-2mm}
\end{equation}
where $\mu$ is a margin-hyperparameter. During inference, all photo-features from the test-gallery ($\{.., f_{p^i}, ..\}$) are pre-computed using the trained encoder $\mathcal{E}_\theta(\cdot)$ and ranked according to their distance from the query-sketch feature ($f_s$). Acc@\textit{q} is then measured as the percentage of sketches retrieving their true-matched photo within the top \textit{q} ranks. For clarity, we dub this as 2D FG-SBIR.

\vspace*{-0mm}
\keypoint{What is wrong with 2D FG-SBIR? }
Being an instance-level matching problem \cite{bhunia2022sketching} FG-SBIR models are generally trained on fixed single-view sketch-photo pairs. Consequently, the naive setup of FG-SBIR as in \cref{equ:triplet_basic} assumes one-to-one correspondence between sketch-photo pairs $(s_i, p_i)$, and typically focuses on shape-matching between them \cite{yu2016sketch}. However as an object in itself is a 3D concept, each instance $i \in {I}$ can be sketched from M$_i$ 2D views as $\mathbf{S}_i = \{s^1_i, \dots, s^{\text{M}_i}_i \}$, corresponding to M$_i$ 2D photos as $\mathbf{P}_i = \{p^1_i, \dots, p^{\text{M}_i}_i \}$ respectively. Now, given a model under naive conditions (\cref{sec:fgsbir_back}) is trained to match sketches from only one fixed-view photo, with its primary focus on shape-matching, the research question arises: `Can such a model's performance generalise to query-sketches drawn from views \textit{different} from its true-match photo in the gallery?'
To answer this, we conduct the following study.

\vspace*{+0.1cm}
\noindent \textbf{Pilot Study: } 
We design a study where we first take a FG-SBIR model pre-trained on fixed single-view sketch-photo pairs of QMUL-ChairV2 dataset~\cite{yu2016sketch} following the basic FG-SBIR training paradigm (\cref{sec:fgsbir_back}) on a VGG-16 \cite{simonyan2015very} backbone. 
Next we curate a test-set using chairs from the dataset by Qi \etal \cite{qi2021FGSBSR}, which has sketches drawn from 0$\degree$, 30$\degree$ and 75$\degree$ and a 3D-shape, per chair-instance. Photos are freely rendered as 24 2D projections (0$\degree$, 15$\degree$, $\cdots$ , 360$\degree$) of the 3D shape. We now evaluate the model on two setups: \textit{(i) `Existing' --} a simple test-set where for every instance, photos \underline{matching} the view of query-sketches (0$\degree$, 30$\degree$ and 75$\degree$) are \textit{present} in the test-gallery along-side other views, and \textit{(ii) `Pilot' --} where they are \textit{absent}. While `\textit{Existing}' scores a satisfactory accuracy (Acc@1) of $58.25$\%, `\textit{Pilot}' drops by $27.15$\% proving that an FG-SBIR model trained on fixed single-view sketch-photo pairs cannot generalise to sketches whose view doesn't match its target photo.

\vspace*{-5mm}
\subsection{Problem Definition}
\label{sec:prob_def}
\vspace*{-2mm}
\noindent Given our findings from the pilot study, we re-examine the problem statement of Fine-Grained Sketch-Based Image Retrieval (FG-SBIR). Currently, FG-SBIR is defined as the task of retrieving the particular target instance from a gallery of photos -- \textit{one} photo per sketch. This definition has lead to state-of-the-art FG-SBIR methods that are mostly \textit{shape-biased} -- use shape matching for retrieval \cite{pang2020solving}. While this definition holds in a 2D setup, in 3D reality, a 3D object can be represented via multiple 2D photos, from different \textit{views} resulting in different sketches for the same instance. Therefore in this work, we for the first time aim to incorporate view-awareness into the FG-SBIR paradigm. Furthermore this paves the way for two new setups where given a sketch, \textit{(i)} retrieve a photo of the instance irrespective of the view in which it is present, \ie, even if its \textit{view does not match} that of the sketch, and \textit{(ii)} retrieve that photo of the instance whose \textit{view exactly matches} that of the sketch.

\keypoint{View-Agnostic FG-SBIR: } 
Given a 2D sketch ($s_{i}^{m}$) of instance $i \in I$, view $m \in [1,\text{M}_i]$, and a gallery of multi-view photos from multiple instances $\{ p_j^m \ | \ \forall \ j \in I , \ m \in [1,\text{M}_i]\}$, we aim to retrieve \textit{any} of all photos $\mathbf{P}_{i} = \{ p_{i}^{1}, \dots, p_{i}^{\text{M}_i} \}$ of the target instance $i \in I$, irrespective of their view.

\keypoint{View-Specific FG-SBIR: } 
Given a 2D sketch ($s_{i}^{m}$) with instance $i \in I$, of view $m \in [1,\text{M}_i]$, and that same gallery of photos, we aim to retrieve that photo of target instance $i \in I$ whose view matches that of the sketch, \ie $p_{i}^{m}$.

\keypoint{Challenges: }
Existing fine-grained datasets like QMUL-ShoeV2 \cite{yu2016sketch}, QMUL-ChairV2~\cite{yu2016sketch} or Sketchy \cite{sangkloy2016sketchy} hold numerous sketch-photo pairs with fine-grained association. However, two major limitations here bottleneck training for view awareness in FG-SBIR: \textit{(i)} They lack any \textit{view-specific} annotations \cite{yu2016sketch} or information, needed to \textit{identify} views. \textit{(ii)} All sketches are drawn from a fixed single-point-of-view matching that of their paired photo, with just one photo \cite{sangkloy2016sketchy} per instance. This lacks the view-\textit{diversity} across \textit{same} instance, needed to instill view-awareness.
We thus aim to learn view-aware knowledge from sketch-\textit{independent} multi-view 2D projections of 3D objects and bridge them across sketch-photo cross-modal dataset to impart cross-modal view awareness into the FG-SBIR \cite{yu2016sketch} pipeline.

\keypoint{Training Dataset: }
To alleviate this dataset issue we need one dataset for learning the cross-modal sketch-photo association ($\mathcal{D}_\text{CM}$) and another containing 2D projections freely rendered from pre-existing 3D shapes ($\mathcal{D}_\text{2D}$). Essentially, $\mathcal{D}_\text{CM} = \{s_i,p_i\}_{i=1}^{\text{N}_\text{CM}}$, which contains $\text{N}_\text{CM}$ sketch ($s$)-photo ($p$) pairs with fine-grained association. Whereas $\mathcal{D}_\text{2D}$ houses $\text{M}_i$ projections for every 3D-shape $\gamma_i$ of $\text{N}_\text{2D}$ shapes, as  
$\mathcal{D}_\text{2D} = \{\{p_i^j\}_{j=1}^{\text{M}_i}\}_{i=1}^{N_\text{2D}}$, where $p_i^j = \mathcal{R}(\gamma_i,{v_j})$. Here, $\{v_j\}_{j=1}^{\text{M}_i}$ refers to the set of select views and $\mathcal{R}(\cdot,\cdot)$ is a 2D-projection rendering from \cite{qi2021FGSBSR}.

\begin{figure}[t]
\centering
\includegraphics[width=\linewidth]{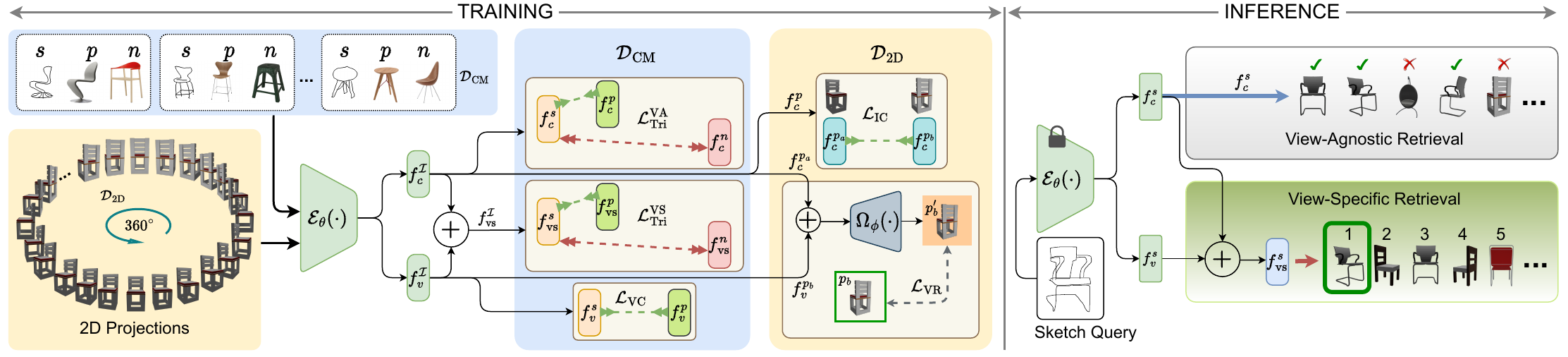}
\vspace*{-6mm}
\caption{
Our model disentangles an input into its \textit{view} and \textit{content} semantics. Sketch-photo pairs from FG-SBIR datasets are used to learn cross-modal discriminative knowledge, whereas multi-view 2D projections from unlabelled 3D models helps condition the encoder with view-aware knowledge. Once trained, the content and view features are used for view-\textit{agnostic} and view-\textit{specific} retrieval as shown.
}
\label{fig:figure2}
\vspace*{-6mm}
\end{figure}

\vspace*{-4mm}
\section{Proposed Methodology}
\vspace*{-3mm}

\keypoint{Overview:}
We aim to devise a framework that learns to incorporate view-awareness in the FG-SBIR training paradigm. Existing literature studying 3D shapes for view-variance usually involves learning a shape descriptor from multi-view images \cite{su2015recognition} or generating complex 3D point-sets from single-view images \cite{psgn}. Being limited by the scarcity of multi-view sketches and also lacking in visual cues compared to an image, limits efficiency of such methods. We thus argue that dealing with the view-variations for cross-modal sketch-photo association effectively, requires a disentanglement model, that explicitly attends to the \textit{view}-semantic. Furthermore, to avoid the complexity of learning 3D-shape descriptors usually followed in parallel shape-retrieval \cite{psgn,su2015recognition} literature, we stick to a 2D-paradigm, given the target is of 2D-image retrieval (not 3D shape).
Accordingly, we aim to design a cross-modal view-aware disentanglement model (\cref{fig:figure2}) that decomposes a photo ($p$) or rasterised sketch ($s$) into a view-semantic part suitable for modelling its `\textit{view}' ($v$) and another component that holds only its \textit{content} ($c$). Such components are trained following a carefully designed learning paradigm (\cref{sec:objectives}) and used  accordingly over a distance metric (\cref{sec:experiments}) for respective view-\textit{agnostic} (VA) or view-\textit{specific} (VS) retrieval.

\keypoint{Model Architecture: } 
To design a view-aware cross-modal encoder $\mathcal{E}_\theta(\cdot)$ that can disentangle the input image into two components -- \textit{view} and \textit{content}, we formally learn the embedding function $\mathcal{E}_\theta: \mathcal{I} \rightarrow \mathbb{R}^d$, that maps an input photo or a rasterized sketch ($\mathcal{I} \in \mathbb{R}^{\text{H}\times\text{W}\times 3}$) to two $d$ dimensional features, where one represents the view of an input image ($f_v^\mathcal{I}$) and the other holds its content ($f_c^\mathcal{I}$), as $f_c^\mathcal{I}, f_v^\mathcal{I} = \mathcal{E}_\theta(\mathcal{I})$.
Our major focus being the training paradigm, we refrain from exploring recent complex backbones of Vision Transformers \cite{wang2021pyramid,dosovitskiy2020image} or Diffusion-based \cite{tang2023emergent,hedlin2023unsupervised} encoders, and employ a simple ImageNet \cite{deng2009imagenet} pretrained VGG-16 network as our backbone feature extractor, followed by an FC-layer for each feature-representation.

\vspace*{-4mm}
\subsection{Learning Objectives} \label{sec:objectives}
\vspace*{-2mm}

\keypoint{Cross-modal Discriminative Learning:}
Being an instance-level matching problem \cite{sain2020cross}, cross-modal discriminatve knowledge is instrumental in training any FG-SBIR framework \cite{bhunia2021more,sain2023exploiting,song2017deep}. Accordingly, we first focus on inducing discriminative knowledge (\cref{sec:fgsbir_back}) to our \textit{content} feature ($f_c^\mathcal{I}$) especially for view-agnostic FG-SBIR, which relies on \textit{content} for cross-modal matching, using sketch-photo pairs from $\mathcal{D}_\text{CM}$. Taking sketch ($s$) as an anchor, we aim to reduce the distance of its \textit{content} ($f_c^s$) from that of its matching photo ($f_c^p$) while increasing it from that of a random non-matching/negative ($n$) photo ($f_c^n$) as:
\vspace*{-1mm}
\begin{equation}
\vspace*{-1mm}
\label{equ:triplet_invariant}
    \mathcal{L}_\text{Tri}^\text{VA} = \max \{0, \mu_c + \delta(f_c^s, f_c^p) - \delta(f_c^s, f_c^n) \} \;.
\end{equation}

However, this alone does not suffice for view-specific retrieval, which additionally needs to distinguish across multiple views of the same instance. 
We thus need to instill view-specific discriminative knowledge as well. Naively imposing triplet-loss \cite{sain2021stylemeup} similarly on $f_v^\mathcal{I}$ alone, would however be sub-optimal as distinguishing amongst \textit{different} views \cite{su2015recognition} is only \textit{relevant} when the model is aware of the associated instances.
We thus combine $f_c^\mathcal{I}$ and $f_v^\mathcal{I}$ over element-wise addition to obtain $f_\text{vs}^\mathcal{I} = f_c^\mathcal{I}+f_v^\mathcal{I}$ which we call the view-specific component ($f_\text{vs}^\mathcal{I}$).
Imposing triplet-loss objective \cite{yu2016sketch} here similarly, we have,
\vspace*{-2mm}
\begin{equation}
\label{equ:triplet_specific}
    \mathcal{L}_\text{Tri}^\text{VS} = \max \{0, \mu_\text{vs} + \delta(f_\text{vs}^s, f_\text{vs}^p) - \delta(f_\text{vs}^s, f_\text{vs}^n) \} \;.
\vspace*{-2mm}
\end{equation}

\keypoint{Learning from 2D projections:}
To instil view awareness in $\mathcal{E}(\cdot)$, it needs to \textit{(i)} recognise the different semantic knowledge coming from different views of the same photo as similar in terms of the content it offers, and \textit{(ii)} discriminate between two different different views as well. As multi-view FG-SBIR sketch-data is rare, rendering most image-based multi-view 3D training paradigms \cite{su2015recognition} sub-optimal, and sketch lacks significantly in visual cues compared to images, rendering single-view 3D reconstruction methods sub-optimal \cite{psgn}, we look to sketch-independent, unpaired 3D shapes \cite{qi2021FGSBSR} to leverage multi-view projections of the same object (in $\mathcal{D}_\text{2D}$) to condition the sketch-photo encoder on view-awareness from photos alone (no sketches involved). 
Accordingly, we introduce two objectives of \textit{(i)} instance-consistency across different projections of the same photo for view-agnostic retrieval, and \textit{(ii)} a cross-view reconstruction objective to further enrich the latent space with cross-view discriminative knowledge. Formally, using our curated dataset of multi-view projections ($\mathcal{D}_\text{2D}$), we pass the set of projections per instance, \ie $\mathbf{P}_i = {p_i^j}|_{j=1}^{\text{M}_i}$ where $p_i^j$ refers to the $j^\text{th}$ view out of $\text{M}_i$ views for the $i^\text{th}$ instance, to extract disentangled content and view-semantics. Attending to \textit{(i)} we take any two out of $\text{M}_i$ views, $p_a$ and $p_b$ (dropping $i$ for brevity) and constrain their \textit{content} ($f_c^{p_{a,b}}$) to ideally occupy the same position in the latent space. Our \textit{instance-consistency} objective thus becomes,
\vspace*{-1mm}
\begin{equation}
\label{equ:instance_match}
    \mathcal{L}_{\text{IC}} = \frac{1}{\binom{\text{M}_i}{2}}\sum_{a=1}^{\text{M}_i-1} \sum_{b=a+1}^{\text{M}_i} \Big|\Big|f_c^{p_{a}} - f_c^{p_{b}}\Big|\Big|_2 \;.
    \vspace*{-1mm}
\end{equation}

For our second objective we advocate for a cross-view translation objective on two different projections ($p_i^a$, $p^b_i$) of the same instance ($p_i$), to enrich the latent space on view-specific knowledge from photos. Specifically, given $\{p_a$, $p_b\}$ and a decoder $\Omega_\phi(\cdot)$ that inputs a d-dimensional feature and outputs an image $\mathcal{I}' \in \mathbb{R}^{H\times W \times 3}$, we perform cross-view reconstruction as $p_b' = \Omega_\phi(f_c^{p_{a}} + f_v^{p_{b}})$. Accordingly, our \textit{view-reconstruction} objective becomes

\vspace*{-3mm}
\begin{equation}
\label{equ:view_rec}
    \mathcal{L}_{\text{VR}} = \frac{1}{\binom{\text{M}_i}{2}}\sum_{a=1}^{\text{M}_i-1} \sum_{b=a+1}^{\text{M}_i} \Big|\Big|p_b' - p_b\Big|\Big|_2 \;.
     \vspace*{-2mm}
\end{equation}

\keypoint{Cross-modal View Consistency:}
Given the cross-modal nature of our task, learning view-awareness only from unlabelled 3D shapes, ignoring sketch, is sub-optimal. Considering that a sketch is paired against only \textit{one} photo in existing FG-SBIR datasets \cite{yu2016sketch,sangkloy2016sketchy} ($\mathcal{D}_\text{CM}$), we thus need to condition $\mathcal{E}_\theta(\cdot)$ towards matching \textit{view} of a sketch ($f_v^s$) with its paired photo ($f_v^p$), to instill cross-modal sketch-photo view-awareness. Although one 
may naively impose triplet-loss for view-consistency on $\{f_v^s,f_v^p,f_v^n\}$ (\cref{equ:triplet_basic}), a major drawback here would be \textit{not} knowing if the view of negative photo ($n$) selected \textit{randomly} is \textit{strictly different} from the positive ($p$) one or \textit{not} (no view-annotations available in $\mathcal{D}_\text{CM}$). This would hence result in a confused guiding signal. Conversely, with the motivation that for a \textit{matching} sketch-photo pair, their \textit{view} should be closer in the latent space, we thus define view-consistency objective as :

\vspace*{-2mm}
\begin{equation}
\label{equ:view_match}
    \mathcal{L}_{\text{VC}} = ||f_v^s - f_v^p||_2 \;.
\end{equation}

\noindent With hyperparameters $\lambda_{1,2}$ our final training  objective is,

\vspace*{-2mm}
\begin{equation}
\label{equ:training}
    \mathcal{L}_\text{trn} = \mathcal{L}_\text{Tri}^\text{VA} + \lambda_1\mathcal{L}_\text{Tri}^\text{VS} +  \lambda_2\left(\mathcal{L}_\text{VC} + \mathcal{L}_\text{IC} + \mathcal{L}_\text{VR}\right) \;.
\end{equation}

\keypoint{Evaluation Paradigm:} Standard FG-SBIR evaluation \cite{yu2016sketch} only focuses on retrieving an instance given a sketch ignoring any dependency on its view, and thus requires only one extracted feature for retrieval. Our motivation of bringing view-awareness in FG-SBIR paradigm splits it as \textit{view-agnostic} and \textit{view-specific} FG-SBIR pipelines which thus requires different feature-types to be used during retrieval. Accordingly, for \textit{view-agnostic} retrieval, the focus being to retrieve the same instance irrespective of the view, we use the \textit{content} feature, $f_c^\mathcal{I}$ of a sketch to match against those of test-gallery photos over a distance-metric. For view-specific retrieval however, we use our combined feature  $f_\text{vs}^\mathcal{I} = f_c^\mathcal{I} + f_v^\mathcal{I}$ similarly, to focus on view as well.

\vspace*{-5mm}
\section{Experiments}\label{sec:experiments}
\vspace*{-4mm}

\keypoint{Datasets:} 
Due to lack of large scale view-incoporated sketch datasets, we rely on our re-purposed fine-grained dataset of $\mathcal{D}_\text{CM}$ + $\mathcal{D}_\text{2D}$ (\cref{sec:fgsbir_back}) for training and evaluation. We focus on two categories -- `chairs' and `lamps', as allowed by the only dataset containing multi-view sketches with fine-grained association by \textbf{Qi \etal} \cite{qi2021FGSBSR}, that houses $555$ and $1005$ sketch/3D-shape quadruplets of `lamps' and `chairs' for $\mathcal{D}_\text{2D}$. Each quadruplet holds three sketches from different views (0$\degree$, 30$\degree$, 75$\degree$ for chairs, and 0$\degree$, 45$\degree$ , 90$\degree$ for lamps) and one 3D-shape. Following \cite{qi2021FGSBSR}, we use $111$ and $201$ quadruplets respectively for testing, and the rest for training. Multi-view photos of the 3D-shapes are freely generated following \cite{qi2021FGSBSR} for [0$\degree$, 30$\degree$, 75$\degree$, 45$\degree$, 90$\degree$, 135$\degree$, 180$\degree$, 225$\degree$, 270$\degree$, 315$\degree$ and 360$\degree$] views for both lamps and chair instances. Importantly, training photos having the \textit{same view} as that of sketches used for inference (0$\degree$, 30$\degree$, 75$\degree$ for chairs and 0$\degree$, 45$\degree$, 90$\degree$ for lamps) are omitted from training-photo set, for a fairer evaluation. Coming to $\mathcal{D}_\text{CM}$, for chairs we use \textbf{QMUL-ChairV2}~\cite{yu2016sketch} containing $1800$/$400$ sketches/photos, entirely for training, unless specified otherwise. As lamps lack any fine-grained sketch-dataset, we curate one from all training-instances of lamps in \cite{qi2021FGSBSR}, where for every lamp we randomly select 1 sketch out of the 3 available (at views 0$\degree$ , 45$\degree$ , 90$\degree$) and pair it with that 2D-projection of its 3D shape, which has the \underline{same} view, thus maintaining the nature of existing FG-SBIR datasets where `sketch-photo pairs are matched against a \underline{fixed} view'. Notably, while inference is performed using the entire \textit{test-set} of $\mathcal{D}_\text{2D}$ (all sketch-views and all 2D projections), sketches of training-set of $\mathcal{D}_\text{2D}$ of chairs and lamps (except those in $\mathcal{D}_\text{CM}^\text{Lamp}$) are intentionally unused to support the motivation of this task.

\keypoint{Implementation Details:}
We use an ImageNet \cite{deng2009imagenet} pre-trained VGG-16~\cite{szegedy2016rethinking} model for $\mathcal{E}(\cdot)$. The decoder ($\Omega$) architecture for cross-view photo reconstruction employs a sequence of stride-2 convolutions with BatchNorm-Relu activation on each convolutional layer except for the output layer where $\operatorname{tanh}$ is used. The encoder's extracted feature is projected into two 128 dimensional vectors -- $f_c^\mathcal{I}$ and $f_v^\mathcal{I}$. We use Adam optimiser with a learning rates of $0.0001$, and batchsize of 64 for 250 epochs. Determining empirically, hyperparameters $\lambda_{1,2}$, $\mu_\text{vs}$ and $\mu_c$ are set to 0.5, 0.7, 0.45 and 0.5 respectively. To reduce the effect of color-bias on retrieval, evident from the uniform color palette evident among projections ($\mathcal{D}_\text{2D}$) of each 3D shape (\cref{fig:figure2}), unlike real photos of $\mathcal{D}_\text{CM}$, we use an off-the-shelf colour augmentation following \cite{lin2021single} on every 2D projection before feeding to the network. Our model was implemented in PyTorch on a 12GB TitanX GPU.

\keypoint{Evaluation Metrics:}
We evaluate both view-agnostic and specific paradigms. As the former can be considered as category-level SBIR evaluation, where each instance denotes a class with its multi-view photos being different instances, we use mean average precision (mAP) \cite{dey2019doodle} and precision for top 100 retrievals (P@100) \cite{dey2019doodle} for evaluation. Whereas, the aim for view-specifc FG-SBIR being to retrieve the photo matching the \textit{exact view} of the query-sketch, we use Acc@q as the percentage of sketches having its true matched photo in the top-$q$ list \cite{song2017deep}.

\vspace*{-3mm}
\subsection{Competitors}
\vspace*{-2mm}

We compare against state-of-the-arts, and a few self-designed baselines by modifying methods from relevant works.
\textbf{\textit{(i)} SoTAs: } \textbf{Triplet-SN}~\cite{yu2016sketch} utilises a Siamese network trained with triplet loss to learn a shared sketch-photo latent space. \textbf{HOLEF-SN}~\cite{song2017deep} improves \cite{yu2016sketch} via spatial attention leveraging a higher order HOLEF ranking loss. \textbf{Triplet-OTF} \cite{bhunia2020onthefly} uses triplet-loss pre-training with RL-based reward maximization for early retrieval. Early retrieval not being our goal, we take its results only on completed sketches. \textbf{StyleVAE} \cite{sain2021stylemeup} employs VAE-based disentanglement via meta-learning for style-agnostic retrieval. \textbf{Jigsaw-CM}~\cite{pang2020solving} uses jigsaw-solving pre-training on mixed photo and edge-map patches, with triplet-based fine-tuning to improve retrieval \textbf{Strong-PVT} \cite{sain2023exploiting} devises a stronger FG-SBIR framework with PVT \cite{wang2021pyramid} backbone -- we use its `Strong' variant. {Notably, `lamps' being an \textit{entirely different} category than the ones these SoTAs were trained on, we do not show SoTA results on lamps.}
\textbf{\textit{(ii)} SoTA++}: Although our method is also trained on ChairV2 \cite{yu2016sketch} for chairs like other SoTAs, reporting SoTAs' evaluation on our curated dataset might be compared to cross-dataset (despite same category: chairs) evaluation. To reduce ambiguity, we reconstruct $\mathcal{D}_\text{CM}$ (only in this setup) for chairs following that for lamps ($\mathcal{D}_\text{CM}^\text{Lamp}$ in \red{\S} Datasets) as $\mathcal{D}_\text{CM}^\text{Chair*}$, and report SoTA results after re-training on $\mathcal{D}_\text{CM}^\text{Chair*}$ and $\mathcal{D}_\text{CM}^\text{Lamp}$.
\textbf{\textit{(iii)} B-Backbones}: Keeping our method same we  explore a few popular architectures used for FG-SBIR as backbone-feature extractor like Inception-V3 \cite{bhunia2020onthefly}, ResNet-50 \cite{he2016deep}, ViT \cite{dosovitskiy2020image} and PVT \cite{wang2021pyramid}.
\textbf{\textit{(iv)} B-Disentangle}: From literature on disentanglement methods \cite{ishfaq2018tvae,lin2018deep,zou2020joint,chowdhury2023scenetrilogy}, we design a few baselines as suggested: \textbf{B-TVAE}~\cite{ishfaq2018tvae} uses a standard VAE with triplet-loss; \textbf{B-DVML}~\cite{lin2018deep} employs a VAE with same-modality translation; 
\textbf{B-Trio} adapts disentanglement module of \cite{chowdhury2023scenetrilogy} to ours.
\textbf{\textit{(v)} B-Misc}: Please note that during training, our setup enforces \textit{no access} to \textit{(a)} multiple sketch-views \textit{(b)} paired 3D-shapes \textit{(c)} cross-modal \textit{association} of one sketch to other views of its target-photo.\ Besides unlabelled 3D-shapes, only 1 sketch-photo pair per instance is available for training. Following \cite{monnier2022share} \textbf{B-Single} ignores the multi-view projections ($\mathcal{D}_\text{2D}$), estimating 3D knowledge from photos and sketches in $\mathcal{D}_\text{CM}$ via single-image 3D reconstruction, independently (separate encoders). During inference it combines its $\{$shape retrieval, texture$\}$ features \cite{monnier2022share} for view-agnostic retrieval whereas $\{$shape, texture, pose$\}$ features for view-specific one. \textbf{B-Pivot} follows \cite{chowdhury2023democratising} using $\mathcal{D}_\text{CM} + \mathcal{D}_\text{2D}$ to design a shared 3D-shape-aware sketch-encoder, to match extracted features from query-sketch and 2D gallery photos for retrieval. \textbf{B-NoProjection} omits using multi-view projections (\ie no $\mathcal{D}_\text{2D}$) keeping the rest same as ours. We also design a two-model baseline (\textbf{B-TwoModel}) with one model per paradigm. For view-agnostic paradigm, we train using $\mathcal{L}_\text{Tri}^\text{VA}$ on sketch-photo triplets ($s$,$p$,$n$) from $\mathcal{D}_\text{CM}$ (\cref{equ:triplet_invariant}), and $\mathcal{L}_\text{IC}$ on 2D projections from $\mathcal{D}_\text{2D}$ (\cref{equ:instance_match}). For view-specific one, we train using $\mathcal{L}_\text{Tri}^\text{VS}$ on similar triplets from $\mathcal{D}_\text{CM}$ (\cref{equ:triplet_specific}), and a simple reconstruction loss ($\mathcal{L}_\text{rec}$=$||\Omega_\phi(p_a)-p_b||$) via our decoder ($\Omega_\phi(\cdot)$) on 2D projections ($p$) from $\mathcal{D}_\text{2D}$.

{\setlength{\tabcolsep}{4.0pt}
\renewcommand{\arraystretch}{0.8}
\begin{table}[t]
\centering
\tiny
\vspace*{-0.2cm}
\caption{Quantitative evaluation for View-Agnostic FG-SBIR}
\vspace*{-4mm}
\input{parts/quant_VA_VS}

\label{tab:quant_VA_VS}
\vspace*{-6mm}
\end{table}
}

\subsection{Performance Analysis}

\vspace*{-4mm}
\keypoint{View-Agnostic Retrieval: }
\Cref{tab:quant_VA_VS} reports quantitative evaluation for view-\textit{agnostic} retrieval.
While \textit{Triplet-SN} \cite{yu2016sketch} and \textit{HOLEF-SN}\cite{song2017deep} score lower, due to their comparatively weaker backbones of Sketch-A-Net \cite{yu2016sketchAnet}, \textit{Jigsaw-CM} \cite{pang2020solving} scores better, given its jigsaw-solving pre-training strategy, enabling better structural knowledge. Although \textit{Triplet-OTF} \cite{bhunia2020onthefly}, with its reinforcement learning-based reward function, surpasses former methods, it is exceeded by \textit{StyleVAE} \cite{sain2021stylemeup} (by 0.07 mAP), thanks to the latter's meta-learning based disentanglement module.
Aided by a much better feature extractor \textit{StrongPVT} \cite{sain2023exploiting} outperforms them all (still 0.12 mAP lower than our PVT \cite{wang2021pyramid}). The overall lower performance of SoTAs, compared to their usual high accuracy \cite{sain2023exploiting}, is likely due to a potential cross-dataset evaluation effect. However their results in SoTA++, obtained by \textit{retraining} SoTAs on $\mathcal{D}_\text{CM}^\text{Chair*, Lamp}$, being coherent with earlier ones, clears shows the demerits of not modelling view-awareness explicitly in FG-SBIR. 
Especially, \textit{StyleVAE++} \cite{sain2021stylemeup}, scores closer to \textit{StrongPVT++} \cite{sain2023exploiting} than earlier, likely due of its ability to disentangle content, based on prior training on disentangling style-invariant features. 
Among other \textit{backbone variants}, PVT \cite{wang2021pyramid} scores best, even better than our initial VGG16-encoder, thanks to its unique pyramidal structure imbibing inductive bias, on feature-maps at multiple-levels. 
While \textit{B-TVAE} and \textit{B-DVML} score lower due to their inferior design, \textit{B-Trio} fares closer thanks to its conditional invertible network. Given our major focus on learning 3D knowledge from 2D data and simplicity of training strategy, the disentanglement module has been kept simple, which can be enhanced further as a future work.
Besides lacking cross-modal discrimination, \textit{B-Single} naively uses 3D-reconstruction objective \cite{monnier2022share} for sketches which is unreliable due to their sparse nature \cite{umar2018abstraction} and lack of visual cues, thus scoring poorly. \textit{B-NoProjection} fares slightly better ($\uparrow$0.095 mAP) with cross-modal discrimination and sketch-photo view-consistency but lags without aid from $\mathcal{D}_\text{2D}$. In contrast, \textit{B-Pivot} \cite{chowdhury2023democratising} excels with well-trained 3D-shape awareness and sketch-photo association. However, lacking any training to

\noindent model \textit{views} explicitly, it lags behind ours.

\keypoint{View-Specific Retrieval: }
From \Cref{tab:quant_VA_VS} we see, the performance trend of different methods reflects similar accuracy shifts, to that seen for view-agnostic retrieval, as in both cases the same trained base encoder model ($\mathcal{E}(\cdot)$) is used for every method, thereby having similar potential for both paradigms. Importantly, unlike its higher performance in the view-agnostic paradigm, \textit{StyleVAE}(++) \cite{sain2021stylemeup} scores much lower, with a larger shift from \textit{StrongPVT}(++) -- likely due to its inability to explicitly attend to the non-\textit{content} part (or \textit{style} in its case) for retrieval. Notably, unlike most methods, ours uses a different feature ($f_\text{vs}^\mathcal{I} = f_c^\mathcal{I} + f_v^\mathcal{I}$) more enriched in view-semantic, thus resulting in better  view-specific retrieval. The low performance of \textit{B-TwoModel} is likely because loss objectives alone are not sufficient to condition the extractor on addressing the `view' component of a sketch, which it needs to disregard (view-agnostic) or emphasise on (view-specific) for target retrieval task, thus justifying our combined paradigm with feature-disentanglement.

\vspace*{-4mm}
\subsection{Ablative Study} 
\vspace*{-3mm}

\keypoint{Importance of loss objectives:} To justify each loss in our framework, we evaluate them in a strip-down fashion (\Cref{tab:abla}), keeping the rest same. FG-SBIR at its core being dependent on cross-modal discrimination, stripping off $\mathcal{L}_\text{Tri}^\text{VS}$ drops Acc@1 significantly (31.15\%). Similarly, stripping $\mathcal{L}_\text{Tri}^\text{VA}$ drops mAP by 0.199. Being the only objective relating \textit{view}-semantic of a sketch with photo, without $\mathcal{L}_\text{VC}$ accuracy dips for both view-\textit{specific} (\textbf{VS}) and \textit{agnostic}, proving its significance. 
\begin{wraptable}[4]{l}{0.58\linewidth}
\setlength{\tabcolsep}{2pt}
\tiny
\centering
\vspace*{-12mm}
\caption{{Ablation of Loss Objectives on `Chairs'}}
\begin{tabular}{lcccccc}
\toprule
Objective-stripped & $\mathcal{L}_\text{Tri}^\text{VA}$ & $\mathcal{L}_\text{Tri}^\text{VS}$ & $\mathcal{L}_\text{VC}$ & $\mathcal{L}_\text{IC}$ & $\mathcal{L}_\text{VR}$ & \cellcolor{YellowGreen!40}Ours-VGG-16\\
\cmidrule(r){1-1} \cmidrule(lr){2-6} \cmidrule{7-7} 
$[$VS$]$ Top-1  (\%) & 25.56 & 21.12 & 55.69 & 52.71 & 56.26 & \cellcolor{YellowGreen!40}60.71 \\
$[$VA$]$ mAP@all     & 0.104 & 0.416 & 0.541 & 0.520 & 0.565 & \cellcolor{YellowGreen!40}0.615 \\
\bottomrule
\end{tabular}
\label{tab:abla}
\end{wraptable}
The need for $\mathcal{L}_\text{IC}$ in view-agnostic paradigm, is evident from the large drop (0.095 mAP) when omitted. A drop of 0.05 mAP and 4.45\% Acc@1 without $\mathcal{L}_\text{VR}$ shows the view-knowledge enrichment it provides in our framework.

\keypoint{Design alternatives: } We explore a few design choices (on chairs) focusing on our loss objectives. \textit{(i)} Given that a sketch relates better to an edgemap than a photo \cite{sain2022sketch3t}, we alter $\mathcal{L}_\text{VR}$ to conduct photo-to-\textit{edgemap} reconstruction of the target view (\ie $p_b'$ in \cref{sec:objectives}). A bit lower score of 57.68\% Acc@1 (0.598 mAP) reveals it to be sub-optimal, likely because, reconstructing a view in the \textit{photo} domain enriches the encoder with other cues like light-intensity \cite{hu2021self}, etc., which is unavailable from edgemaps. 
(ii) Modifying $\mathcal{L}_\text{VC}$ as a triplet loss  on $\{f_v^s$, $f_v^p$, $f_v^n\}$ using \cref{equ:triplet_basic} dips accuracy, especially for view-specific paradigm (by 4.5\% Acc@1) as during training we are unaware (no annotations), if the negative's ($n$) and positive's views are strictly \textit{different} or \textit{not}, thus creating a sub-optimal gradient for encoder update.
(iii) Omitting colour augmentation during model training (chairs) invokes a colour bias \cite{lin2021single}, dropping accuracy by 0.065 mAP (2.85\% Acc@1), thus proving its importance.
(iv) Utilising separate VGG-16 encoders extracting `content' ($f_c^\mathcal{I}$) and `view' ($f_v^\mathcal{I}$) features yields poor results of $0.528$ mAP@all on view-agnostic and $54.32\%$ Top-1 score on view-specific FG-SBIR -- likely because using different extractors yields poor coherence between $f_v^\mathcal{I}$ and $f_c^\mathcal{I}$, as unlike \textit{ours}, they do not implicitly condition the model on the knowledge that both content and view features belong to the same instance. This is \textit{crucial} for \textit{view-awareness}, especially for view-specific feature's representation ($f_\text{vs}^\mathcal{I}$) which combines both features for subsequent training and retrieval.

\keypoint{Performance under low-data regime}: Our framework being manoeuvred to deal with data-scarcity of sketch-views (\ie not much data to learn view-awareness within sketches sketch-view learning ), we aim to explore our generalisation potential under low-data regime. Accordingly we vary training data ($\mathcal{D}_\text{CM}$) 
\begin{wrapfigure}[8]{l}{0.5\linewidth}
\centering
    \vspace*{-9mm}
    \includegraphics[width=\linewidth]{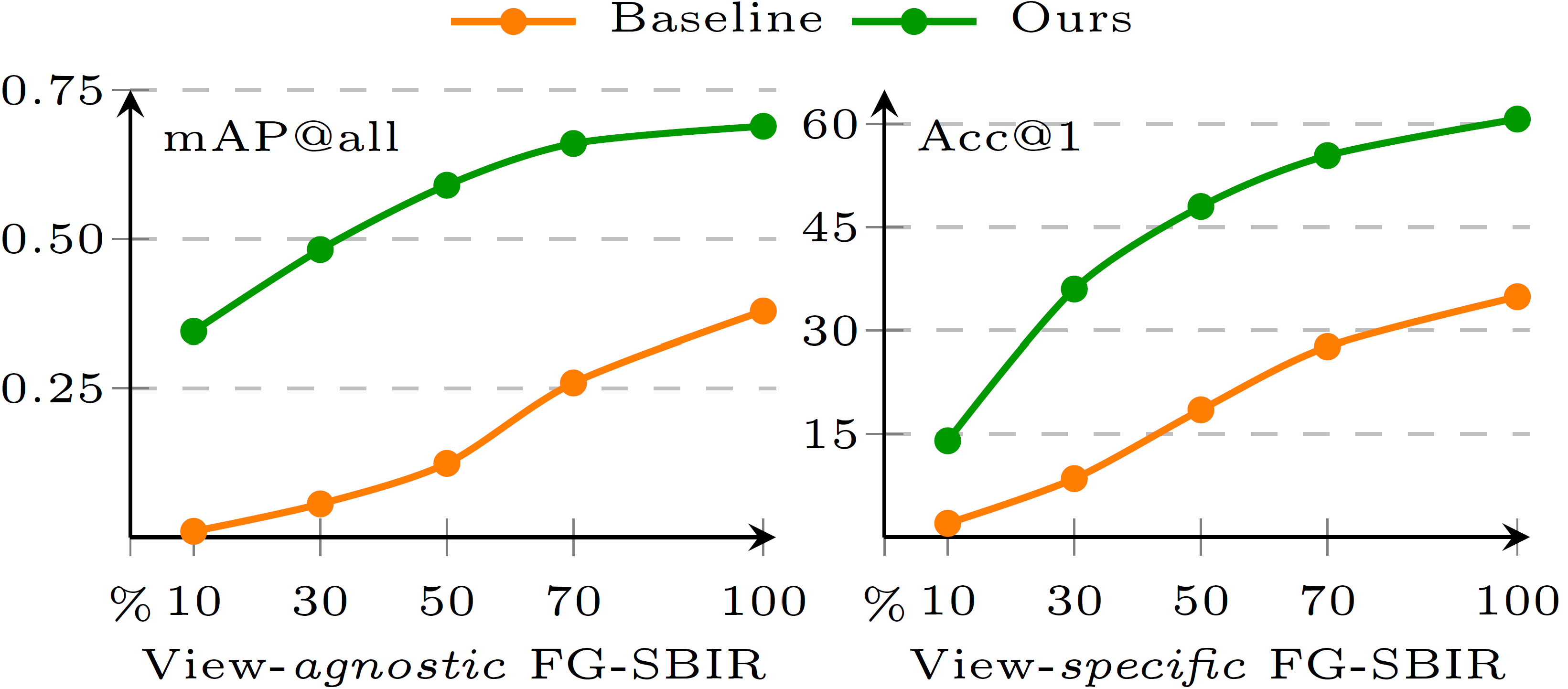}
    \vspace*{-7mm}
    \caption{Varying training data-size ($\mathcal{D}_\text{CM}$).}
    \label{fig: generalisation}
    \vspace*{-7mm}
\end{wrapfigure}
for chairs as 10\%, 30\%, 50\%, 70\% and 100\%. Our view-agnostic and view-specific FG-SBIR \cite{song2017deep} performances remain relatively stable (\cref{fig: generalisation}) at variable training-data-size, compared to a baseline of Triplet-SN -- thanks to our carefully designed objectives (\cref{sec:objectives}), especially those of $\mathcal{L}_\text{IC}$ and $\mathcal{L}_\text{VR}$ for additionally enriching the latent space with cross-view photo knowledge, apart from cross-modal triplet loss \cite{yu2016sketch}.

\keypoint{Further Insights: }
\textit{(i)} Optimal feature dimension for both content and view features were empirically found to be 128, with stable results at higher ones. \textit{(ii)} \textit{Our-VGG16} utilises $14.71$ mil. params with {\small$\sim$} $40.18$G FLOPs and takes 0.16ms (0.21ms) for view-sepcifc (agnostic) retrieval per query during evaluation -- close to 0.18ms of \textit{Triplet-SN}.
\textit{(iii)} On varying each of $\lambda_{1,2}$, as $\{0.1, 0.15, \cdots, 0.9\}$ independently, accuracy falls when: \textit{(a)} $\lambda_1$ $>$ 0.55 or $\lambda_2$ $<$ 0.65 and (b) $|\lambda_1-\lambda_2|$ is large, (\eg $\lambda_{1,2}$ = $0.2$, $0.8$), giving us optimal values at $\lambda_{1,2}$ = $0.5$, $0.7$ empirically. Following FG-SBIR works on margin hyperparameters \cite{yu2016sketch,song2017deep} $\mu_\text{vs}$ and $\mu_\text{c}$ were varied as $\{0.3, 0.35, \cdots, 0.7\}$, delivering optimal values at $\mu_\text{vs}$ = $0.45$, $\mu_\text{c}$ = $0.5$.

\vspace*{-4mm}
\section{Limitations and Future Works} 
\vspace*{-3mm} 
\textit{(i)} Besides \textit{off-the-shelf} complex feature-extractors  (\Cref{tab:quant_VA_VS}), future works may explore designing sketch-specific modules or complex architectures like DINOV2 \cite{oquab2023dinov2} for view-aware feature extraction to enhance accuracy.
\textit{(ii)} Other alternatives for disentanglement paradigms based on meta-learning \cite{sain2021stylemeup} or diffusion \cite{kim2022diffusionclip} could be explored.
(\textit{iii}) Alleviating data scarcity for sketch-views might allow recent methods \cite{xu20233difftection} that learn cross-modal 3D knowledge from multi-view images, to enhance robustness of view-aware FG-SBIR paradigms.

\vspace*{-4mm}
\section{Conclusion}
\vspace*{-3mm}
In this paper we propose a system that addresses the nuanced challenge of view selection in FG-SBIR, seamlessly accommodating both view-agnostic and view-specific retrieval approaches. The introduction of multi-view 2D rendered projections of 3D objects aims to overcome dataset limitations, promoting cross-modal view awareness within the FG-SBIR pipeline. Additionally, our implementation of a customisable cross-modal feature, facilitated by a disentanglement framework, allows users to fluidly transition between view-agnostic and view-specific retrieval modes, enhancing system adaptability and user experience.

\input{arx_supp}
\newpage
\bibliographystyle{splncs04}
\bibliography{main}
\end{document}

%% file: arx_preamble.tex
\usepackage{graphicx}
\usepackage{amsmath}
\usepackage{amssymb}
\usepackage{booktabs}
\usepackage{color}
\usepackage{amsthm}
\usepackage[dvipsnames]{xcolor}
\usepackage{multirow}
\usepackage{mathtools}
\usepackage{soul}
\usepackage{nicefrac}
\usepackage{stmaryrd}
\usepackage[innercaption]{sidecap}
\usepackage{colortbl}
\usepackage[accsupp]{axessibility}
\usepackage{array, makecell} 
\usepackage{wrapfig}

\def\red#1{\textcolor[rgb]{1,0,0}{#1}}

\definecolor{newcolor}{RGB}{133,99,99}

\newcommand{\keypoint}[1]{\vspace{0.1cm}\noindent\textbf{#1}\;}
\newcommand{\cut}[1]{}

\newcolumntype{g}{>{\columncolor{YellowGreen!40}}c}

\newcommand{\MYhref}[3][blue]{\href{#2}{\color{#1}{#3}}}
\definecolor{deepGreen}{RGB}{0,153,0}
\definecolor{orange}{RGB}{255,125,0}
\definecolor{Gray}{gray}{0.9}

\usepackage{gensymb}
\usepackage{textcomp}

%% file: parts/quant_VA_VS.tex
\begin{tabular}{cr|cccc|cccc}
\toprule
\multicolumn{2}{c}{\multirow{5}{*}{Methods}} & \multicolumn{4}{|c|}{View-Agnostic} & \multicolumn{4}{c}{View-Specific}\\
\cmidrule(lr){3-6} \cmidrule(l){7-10}
& & \multicolumn{2}{c}{Chairs} & \multicolumn{2}{c|}{Lamps} & \multicolumn{2}{c}{Chairs} & \multicolumn{2}{c}{Lamps}\\
\cmidrule(lr){3-4} \cmidrule(l){5-6} \cmidrule(lr){7-8} \cmidrule(l){9-10}
& & mAP@all & P@100 & mAP@all & P@100 & Top-1 & Top-10 & Top-1 & Top-10\\
\cmidrule(r){1-2} \cmidrule(lr){3-4} \cmidrule(lr){5-6} \cmidrule(lr){7-8} \cmidrule(l){9-10}
\multirow{6}{*}{\rotatebox{0}{SoTA}}
& Triplet-SN \cite{yu2016sketch}        & 0.379 & 0.447 & -- & -- & 34.88 & 76.62 & -- & --\\
& HOLEF-SN \cite{song2017deep}          & 0.398 & 0.454 & -- & -- & 37.23 & 78.63 & -- & --\\
& Jigsaw-CM \cite{pang2020solving}      & 0.432 & 0.525 & -- & -- & 41.14 & 81.78 & -- & --\\
& Triplet-OTF \cite{bhunia2020onthefly} & 0.447 & 0.514 & -- & -- & 42.21 & 82.79 & -- & --\\
& StyleVAE \cite{sain2021stylemeup}     & 0.523 & 0.602 & -- & -- & 46.19 & 87.66 & -- & --\\
& StrongPVT \cite{sain2023exploiting}   & 0.569 & 0.624 & -- & -- & 55.93 & 90.78 & -- & --\\
\cmidrule(r){1-2} \cmidrule(lr){3-4} \cmidrule(lr){5-6} \cmidrule(lr){7-8} \cmidrule(l){9-10}
\multirow{5}{*}{\makecell[c]{B-Backbones\\(Ours)}}
& ViT \cite{dosovitskiy2020image}        & 0.385 & 0.415 & 0.338 & 0.399 & 34.38 & 76.18 & 33.96 & 75.53\\
& ResNet-50 \cite{he2016deep}            & 0.451 & 0.536 & 0.415 & 0.511 & 47.15 & 88.02 & 46.21 & 87.11\\
& Inception-V3 \cite{bhunia2020onthefly} & 0.512 & 0.573 & 0.468 & 0.542 & 50.18 & 90.19 & 49.11 & 89.23\\
& \cellcolor{YellowGreen!40}VGG-16 \cite{simonyan2015very} & \cellcolor{YellowGreen!40}0.615 & \cellcolor{YellowGreen!40}0.693 & \cellcolor{YellowGreen!40}0.552 & \cellcolor{YellowGreen!40}0.664 \cellcolor{YellowGreen!40}& \cellcolor{YellowGreen!40}60.71 & \cellcolor{YellowGreen!40}91.18 & \cellcolor{YellowGreen!40}60.56 & \cellcolor{YellowGreen!40}90.62\\
\rowcolor{YellowGreen!40} \cellcolor{white}
& PVT \cite{wang2021pyramid}            & 0.689 & 0.742 & 0.628 &0.716 & 67.11 & 91.78 & 65.35 &  92.97 \\
\cmidrule(r){1-2} \cmidrule(lr){3-4} \cmidrule(lr){5-6} \cmidrule(lr){7-8} \cmidrule(l){9-10}
\multirow{3}{*}{\makecell[c]{B-Disentangle}}
& B-TVAE \cite{ishfaq2018tvae}              & 0.394 & 0.449 & 0.345 & 0.414 & 36.89 & 77.91 & 35.28 & 74.92 \\
& B-DVML \cite{lin2018deep}                 & 0.417 & 0.478 & 0.381 & 0.458 & 45.63 & 86.94 & 43.21 & 84.11 \\
& B-Trio \cite{chowdhury2023scenetrilogy}   & 0.572 & 0.629 & 0.501 & 0.582 & 58.68 & 90.85 & 55.63 & 89.02 \\
\cmidrule(r){1-2} \cmidrule(lr){3-4} \cmidrule(l){5-6}
\cmidrule(r){1-2} \cmidrule(lr){3-4} \cmidrule(l){5-6} \cmidrule(lr){7-8} \cmidrule(l){9-10}
\multirow{4}{*}{\makecell[c]{B-Misc}}
& B-Single \cite{monnier2022share}            & 0.221 & 0.281 & 0.184 & 0.233 & 18.68 & 45.68 & 17.91 & 44.69\\
& B-Pivot \cite{chowdhury2023democratising}   & 0.316 & 0.401 & 0.295 & 0.362 & 55.92 & 90.62 & 53.62 & 88.65\\
& B-TwoModel                                  & 0.421 & 0.498 & 0.382 & 0.459 & 48.23 & 89.21 & 46.93 & 87.75\\
& B-NoProjection                              & 0.592 & 0.667 & 0.529 & 0.611 & 50.79 & 89.93 & 48.73 & 87.98\\

\cmidrule(r){1-2} \cmidrule(lr){3-4} \cmidrule(lr){5-6} \cmidrule(lr){7-8} \cmidrule(l){9-10}
\morecmidrules \cmidrule(r){1-2} \cmidrule(lr){3-4} \cmidrule(lr){5-6} \cmidrule(lr){7-8} \cmidrule(l){9-10}
\multirow{7}{*}{\rotatebox{0}{\makecell[c]{SoTA++\\$(\mathcal{D}_\text{CM}^\text{Chair*})$}}}
& Triplet-SN \cite{yu2016sketch}        & 0.416 & 0.476 & 0.378 & 0.451 & 43.09	& 83.29 & 41.32 & 81.48\\
& HOLEF-SN \cite{song2017deep}          & 0.428 & 0.502 & 0.387 & 0.466 & 45.78	& 87.33 & 43.89 & 85.42\\
& Jigsaw-CM \cite{pang2020solving}      & 0.492 & 0.539 & 0.442 & 0.518 & 48.51	& 88.59 & 46.51 & 86.67\\
& Triplet-OTF \cite{bhunia2020onthefly} & 0.521 & 0.591 & 0.476 & 0.571 & 49.53	& 89.66 & 47.49 & 87.71\\
& StyleVAE \cite{sain2021stylemeup}     & 0.618 & 0.675 & 0.553 & 0.644 & 54.36 & 90.71 & 52.12 & 88.73\\
& StrongPVT \cite{sain2023exploiting}   & 0.641 & 0.708 & 0.584 & 0.677 & 64.68	& 91.15 & 62.02 & 90.15\\
\rowcolor{YellowGreen!40} \cellcolor{white}
& Ours-PVT \cite{wang2021pyramid}       & 0.702 & 0.771 & 0.681 & 0.749 & 70.26 & 92.86 & 68.32 & 93.04\\
\bottomrule
\end{tabular}

%% file: arx_supp.tex
{\onecolumn{
\begin{center}
\large{\textbf{Supplementary material for \\Freeview Sketching: View-Aware Fine-Grained Sketch-Based Image Retrieval}}
\end{center}
}

\section*{A. Qualitative results of View Aware FG-SBIR}
\vspace{-1mm}
\cref{fig:quals_VA} shows qualitative comparison for View-\textit{Agnostic} FG-SBIR, of a baseline method (\textit{Triplet-SN}) \textit{vs} ours (\textit{Ours-VGG-16}), on our standard train-test setting (Sec.\red{5} -- Datasets). \cref{fig:quals_VS} illustrates the same for View-\textit{Specific} FG-SBIR, where continuous green rectangles denote the \textit{target view-matched} photo and the dashed ones depict other \textit{views} of the \textit{target instance}, just for clarity. Numbers in boxes (green/white) represent corresponding ranks of retrieved photos.

\begin{figure}[!hbt]
    \centering
    \vspace{-4mm}
    \includegraphics[width=\linewidth,trim={0 1mm 0 0},clip]{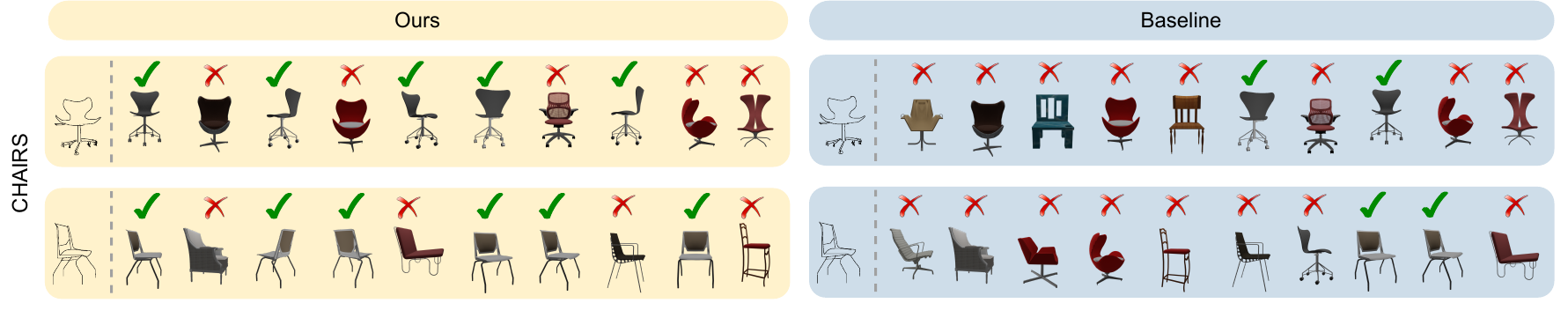}
    \includegraphics[width=\linewidth]{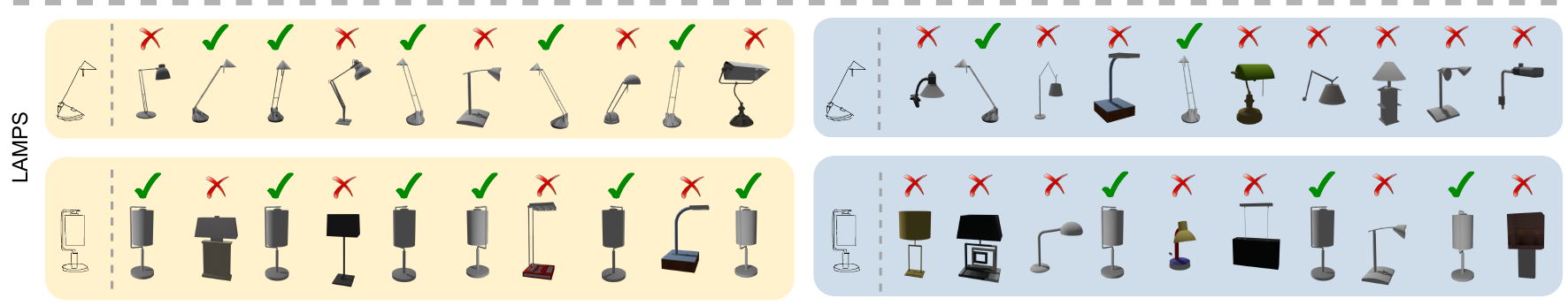} 
    \vspace{-7mm}
    \caption{Qualitative Results of View-Agnostic FG-SBIR}
    \label{fig:quals_VA}
    \vspace{-6mm}
\end{figure}

\begin{figure}[!hbt]
    \centering
    \vspace{-4mm}
    \includegraphics[width=\linewidth]{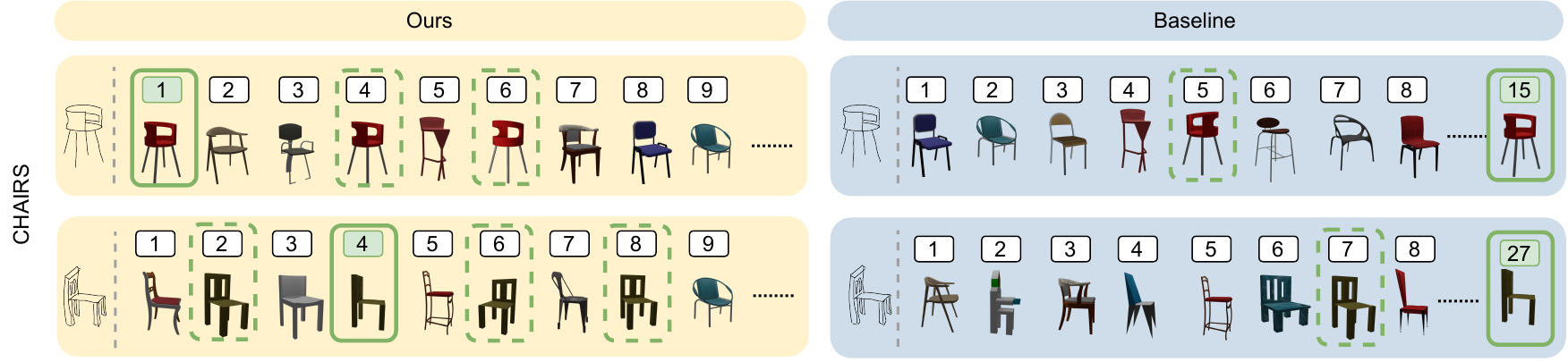}
    \includegraphics[width=\linewidth]{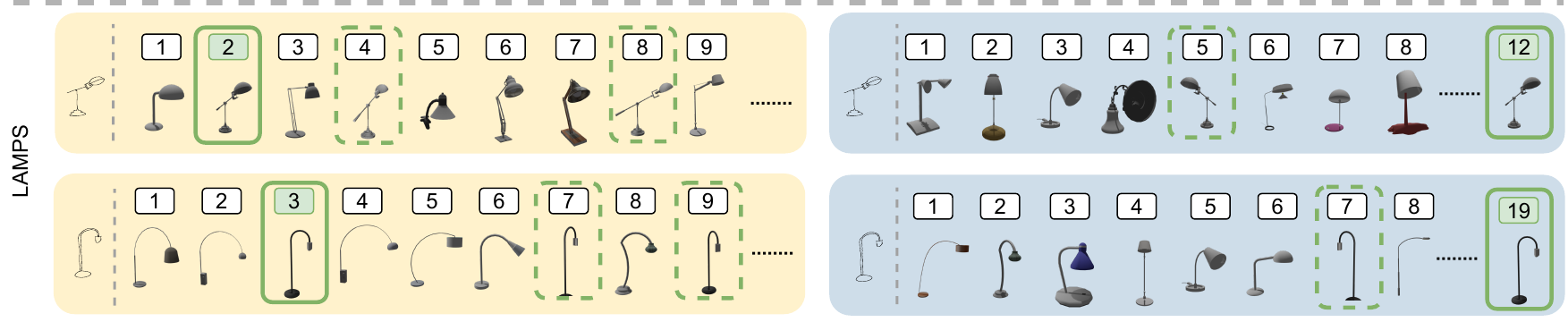}
    \vspace{-7mm}
    \caption{Qualitative Results of View-Specific FG-SBIR}
    \label{fig:quals_VS}
    \vspace{-4mm}
\end{figure}

\newpage
\section*{B. Details on Two-model baseline}

Following figure details the construction of the two-model baseline described under Section \red{5.1} as \textbf{B-TwoModel}. Here we design one model for each paradigm. For each model, a backbone feature-extractor $\mathcal{E}(\cdot)$ (VGG-16) extracts a \textit{single} d-dimensional feature \textit{without} disentanglement.
For \textbf{view-agnostic} paradigm (Please see \cref{fig:Twomodel} left), we train using $\mathcal{L}_\text{Tri}^\text{VA}$ on sketch-photo triplets ($s$,$p$,$n$) from $\mathcal{D}_\text{CM}$ (Eq. (\red{2})), and $\mathcal{L}_\text{IC}$ on 2D projections from $\mathcal{D}_\text{2D}$ (Eq. (\red{4})). 
For \textbf{view-specific} one (Please see \cref{fig:Twomodel} right), we train using $\mathcal{L}_\text{Tri}^\text{VS}$ on similar triplets from $\mathcal{D}_\text{CM}$ (Eq. (\red{3})), and a simple reconstruction loss ($\mathcal{L}_\text{rec}$ = $||\Omega_\phi(p_a)-p_b||$) via our decoder ($\Omega_\phi(\cdot)$) on 2D projections ($p$) from $\mathcal{D}_\text{2D}$.

\begin{figure}[!hbt]
    \centering
    \includegraphics[width=\linewidth]{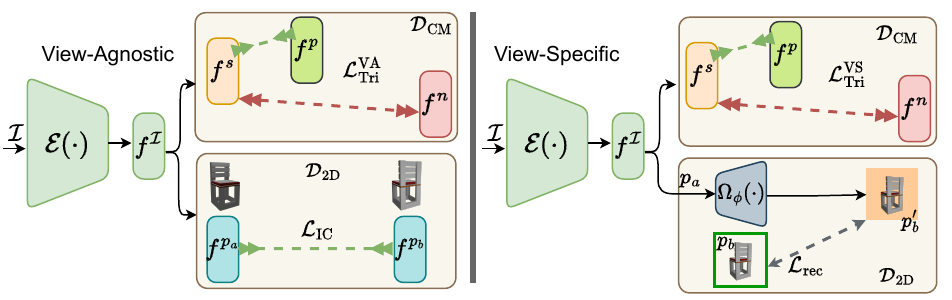}
    \caption{B-Twomodel framework. Left: Model for View-Agnostic paradigm. Right: Model for View-Specific paradigm.}
    \label{fig:Twomodel}
    \vspace{-4mm}
\end{figure}

While the view-agnostic model scores $0.421$ and $0.382$ mAP (\vs $0.615$ and $0.552$ mAP of \textit{Ours-VGG-16}) on Chairs and Lamps respectively, the view-specific one scores $48.23$\% and $46.93$\% Top-1 accuracy (\vs $60.71$\% and $60.56$\% of Ours-VGG16) on Chairs and Lamps respectively (Please see Table \red{1} in main paper). This shows that while being dedicated to one task, having a separate model might seem to tackle each problem better, unfortunately the loss objectives alone are not sufficient to condition the extractor on addressing the ‘view’ component of a sketch, which it needs to \textit{disregard} (view-agnostic) or \textit{emphasise} on (view-specific) for target retrieval task, thus justifying our combined paradigm with feature-disentanglement.

}